\pdfoutput=1

\documentclass[11pt, twocolumn]{article}

\usepackage[]{ACL2023}

\usepackage{times}
\usepackage{latexsym}
\usepackage{hyperref}
\usepackage{graphicx}
\usepackage{verbatim}
\usepackage{booktabs}
\usepackage{float}

\usepackage[T1]{fontenc}

\usepackage[utf8]{inputenc}

\usepackage{microtype}

\usepackage{inconsolata}

\newcommand{\name}[1][]{\textsc{CheXpert Plus}}

\usepackage{multicol} 

\usepackage{inconsolata}
\usepackage{listings}
\usepackage{color}
\usepackage{algorithm} 
\usepackage{algpseudocode} 
\usepackage{graphicx} 
\usepackage{float}    
\usepackage{pifont}
\usepackage{booktabs} 
\usepackage{tabularx} 
\usepackage{geometry} 
\usepackage{amsmath} 

\usepackage{times}
\usepackage{microtype}
\usepackage{epsfig}
\usepackage{caption}
\usepackage{float}
\usepackage{placeins}
\usepackage{color, colortbl}
\usepackage{stfloats}
\usepackage{enumitem}
\usepackage{tabularx}
\usepackage{xstring}
\usepackage{multirow}
\usepackage{xspace}
\usepackage{url}
\usepackage{subcaption}
\usepackage{xcolor}
\usepackage[hang,flushmargin]{footmisc}

\definecolor{mylightgray}{gray}{0.9}

\definecolor{codegreen}{rgb}{0,0.6,0}
\definecolor{codegray}{rgb}{0.5,0.5,0.5}
\definecolor{codepurple}{rgb}{0.58,0,0.82}
\definecolor{backcolour}{rgb}{0.95,0.95,0.92}
\definecolor{mypurple}{RGB}{200,192,248}
\definecolor{mypurpledeep}{RGB}{142,126,240}
\definecolor{mygreen}{RGB}{117,170,156}
\definecolor{myyellow}{RGB}{255,192,0}
\definecolor{myblue}{RGB}{57,143,255}
\definecolor{mygrey}{RGB}{231,230,230}
\definecolor{codey}{RGB}{220,220,170}
\definecolor{coder}{RGB}{206,145,120}
\definecolor{codeb}{RGB}{156,220,254}
\definecolor{codenum}{RGB}{204,204,204}

\DeclareSymbolFont{extraup}{U}{zavm}{m}{n}
\DeclareMathSymbol{\varheart}{\mathalpha}{extraup}{86}
\DeclareMathSymbol{\vardiamond}{\mathalpha}{extraup}{87}

\lstdefinestyle{mystyle}{
    backgroundcolor=\color{backcolour},   
    commentstyle=\color{codegreen},
    keywordstyle=\color{magenta},
    numberstyle=\tiny\color{codegray},
    stringstyle=\color{codepurple},
    basicstyle=\footnotesize,
    breakatwhitespace=false,         
    breaklines=true,                 
    captionpos=b,                    
    keepspaces=true,                 
    numbers=left,                    
    numbersep=5pt,                  
    showspaces=false,                
    showstringspaces=false,
    showtabs=false,                  
    tabsize=2
}

\newcommand{\thankswithlink}[1]{%
  \begingroup
  \renewcommand\thefootnote{}
  \footnote{\href{#1}{#1}}
  \addtocounter{footnote}{-1}
  \endgroup
}

\title{MedSlice: Fine-Tuned Large Language Models \\ for Secure Clinical Note Sectioning}

\author{
    Joshua Davis$^{1,2*}$ \\
    \texttt{joshua\_davis@dfci.harvard.edu}
    \\\And
    Thomas Sounack$^{1*}$\\
    \texttt{thomas\_sounack@dfci.harvard.edu}
    \\\AND
    Kate Sciacca$^{1,3}$\\
    \\\And
    Jessie M Brain$^{1,3}$\\
    \\\And
    Brigitte N Durieux$^{1,4}$\\
    \\\AND
    Nicole D Agaronnik$^{1,5}$\\
    \\\And
    Charlotta Lindvall$^{1,3,5}$\\
    \affil{
       $^1$ Dana-Farber Cancer Institute \hspace{0.2cm}
       $^2$ Albany Medical College \hspace{0.2cm}
       $^3$ Brigham and Women’s Hospital \hspace{0.2cm}\\
       $^4$ McGill University \hspace{0.5cm}
       $^5$ Harvard Medical School
    }
}

\begin{document}

\maketitle

\thankswithlink{https://github.com/lindvalllab/MedSlice}

\begin{abstract}

\textbf{Objective} \hspace{1.5mm} Extracting sections from clinical notes is crucial for downstream analysis but is challenging due to variability in formatting and labor-intensive nature of manual sectioning. While proprietary large language models (LLMs) have shown promise, privacy concerns limit their accessibility. This study develops a pipeline for automated note sectioning using open-source LLMs, focusing on three sections: History of Present Illness, Interval History, and Assessment and Plan.

\textbf{Materials and Methods} \hspace{1.5mm} We fine-tuned three open-source LLMs to extract sections using a curated dataset of 487 progress notes, comparing results relative to proprietary models (GPT-4o, GPT-4o mini). Internal and external validity were assessed via precision, recall and F1 score.

\textbf{Results} \hspace{1.5mm} Fine-tuned Llama 3.1 8B outperformed GPT-4o ($F1=0.92$). On the external validity test set, performance remained high ($F1= 0.85$).

\textbf{Discussion and Conclusion} \hspace{1.5mm} Fine-tuned open-source LLMs can surpass proprietary models in clinical note sectioning, offering advantages in cost, performance, and accessibility.

\end{abstract}

\section{Background And Significance}

Clinical documentation is critical for patient care, facilitating communication across clinicians and providing a comprehensive record of patient progress from inpatient to outpatient settings. While clinical notes often follow semi-structured formats, such as SOAP or sectioned templates (e.g., History of Present Illness, Family History, Review of Systems, Physical Exam, Assessment, and Plan), they also contain rich, unstructured free-text narratives documenting a clinician’s direct observations and assessments \citep{Podder2023}. Though unstructured/semi-structured free text contains valuable clinical information, the variability in formatting between individual documenting clinicians presents a challenge in the research setting. Manually “sectioning” of notes to find current information is labor-intensive, error-prone, and unsuitable for large-scale data analysis \citep{Sheikhalishahi2019}. Prior efforts to automate this process have included rule-based heuristics and machine learning models \citep{Denny2009, Eyre2022, Pomares-Quimbaya2019}; however, these approaches have limited generalizability across diverse note types, hospital systems, and clinical domains.\\

The emergence of large language models (LLMs) presents a transformative opportunity for section segmentation in clinical documentation \citep{Zhou2024}. Unlike earlier approaches, LLMs are trained on diverse datasets, enhancing their adaptability to varied formats and institutions \citep{article_7}. Successful implementation of these methods could enable streamlined workflows, focusing on extracting and analyzing specific sections of interest from clinical notes. A previous study found that proprietary LLMs, such as OpenAI's GPT-4, achieved an average F1 score of 0.77 in identifying note sections \citep{Zhou2024}. While this represents a promising initial result, access to these models is often limited due to privacy concerns. This study also tested open-source models but reached a lower performance than GPT-4. Our work implements a similar methodology, but focuses on specific sections of interest and a curated dataset to achieve state-of-the-art performance on this task with smaller fine-tuned LLMs (<8 billion parameters). We test for robustness using data from various cancer centers and institutions. By optimizing smaller models for targeted domains, such as History of Present Illness, Interval History, and Assessment and Plan, we aim to create accessible methods that improve efficiency in extracting sections of interest from clinical notes for downstream analysis.
\section{Objective}

This study aims to develop an automated method to extract clinically relevant sections of notes essential for downstream analysis, using a scalable pipeline compatible with local and cloud hardware.
\section{Materials and Methods}

\subsection{Dataset}

\begin{table*}[hb!]
\centering
\resizebox{0.9\textwidth}{!}{%
\begin{tabular}{l c c c c} 
\toprule[1.5pt]
\multicolumn{1}{c}{\cellcolor[HTML]{EFEFEF}}
& \multicolumn{1}{c}{\cellcolor[HTML]{EFEFEF}All Notes}
& \multicolumn{1}{c}{\cellcolor[HTML]{EFEFEF}Breast}
& \multicolumn{1}{c}{\cellcolor[HTML]{EFEFEF}GI}
& \multicolumn{1}{c}{\cellcolor[HTML]{EFEFEF}Neuro}
\\ 
\midrule
\addlinespace
\# Notes
& 1,147
& 487
& 465
& 195
\\
\midrule
\# Unique patients
& 433
& 157
& 254
& 22
\\
\midrule
Provider (\%)
& 
&
&
&
\\
\hspace{.5 cm} \textit{Physician}
& 61.7
& 68.0
& 59.8
& 50.8
\\
\hspace{.5 cm} \textit{Nurse Practitioner}
& 29.7
& 25.3
& 29.0
& 42.6
\\
\hspace{.5 cm} \textit{Physician Assistant}
& 8.5
& 6.8
& 11.2
& 6.7
\\
\midrule
Average \# of tokens (95\% CI)
& \begin{tabular}[c]{@{}c@{}}1,814 \\ (1,737 - 1,891) \end{tabular}
& \begin{tabular}[c]{@{}c@{}}1,789 \\ (1,671 - 1,907) \end{tabular}
& \begin{tabular}[c]{@{}c@{}}1,942 \\ (1,813 - 2,071) \end{tabular}
& \begin{tabular}[c]{@{}c@{}}1,570 \\ (1,737 - 1,726) \end{tabular}
\\
\midrule
Notes containing (\%)
&
&
&
&
\\
\hspace{.5 cm} \textit{Recent Clinical History}
& 86.6
& 86.0
& 92.5
& 73.8
\\
\hspace{.5 cm} \textit{Assessment and Plan}
& 87.2
& 87.3
& 92.5
& 74.4
\\
\bottomrule[1.5pt]
\end{tabular}
} 
\caption{Description of the dataset}
\label{table-dataset}
\end{table*}

Clinical notes from three oncology groups (breast, gastrointestinal, neurological) were annotated by two nurse practitioners (KS and JB). The first 25 notes from the gastrointestinal group were independently coded to facilitate initial data familiarization and the development of a codebook. Using this preliminary codebook, KS and JB independently coded a total of 653 notes, identifying spans related to the history of present illness, interval history, and assessment \& plan (A\&P). Due to variability in documentation, the history of present illness and interval history were combined into a single label, recent clinical history (RCH).\\

Inter-rater reliability was calculated using Jaccard Index (JI) \citep{article_7}. For sections where the JI between the two annotations exceeded 80\%, the union of the annotations was adopted as the final label. A total of 125 notes did not meet this threshold and were re-coded through group discussion involving all annotators and a third-party adjudicator (JD). This process resulted in the finalized codebook (\autoref{Codebook}). An additional 494 notes were single coded by KS using the finalized codebook, culminating in a dataset of 1,147 clinical notes (\autoref{table-dataset}).

\subsection{Baseline}

For baseline evaluation, we tested two rule-based approaches: SecTag and the sectioner module from MedSpaCy \citep{Eyre2022, Denny2009}. SecTag employs terminology-based rules and naive Bayesian scoring to identify section headers in clinical notes, while MedSpaCy, an updated version of SecTag used by the VA in multiple studies \citep{chapman-etal-2020-natural, Chapman2021}, builds upon this methodology. Both tools were adapted for compatibility with our processing pipeline.\\

In addition to these baselines, we utilized a Clinical-Longformer with a 4096-token context window \citep{Li2023}, trained with a custom head to predict the start and end positions of target sequences. Using a dataset of 487 notes from the breast cancer center, we trained two separate models: one for extracting RCH and another for A\&P.

\subsection{Models}

Five LLMs (GPT-4o, GPT-4o mini \citep{openai2024gpt4ocard}, Llama 3.2 instruct (1B), Llama 3.2 instruct (3B), Llama 3.1 instruct (8B) \citep{grattafiori2024llama3herdmodels}) were evaluated for section identification. OpenAI models ran on a HIPAA-compliant endpoint \citep{Umeton2023}, while Meta models were run on a virtual machine with a context window of 8192 tokens. All used a unified prompt (\autoref{prompt}); OpenAI models applied function-calling, and Meta models were tested pre and post supervised fine-tuning (SFT) \citep{wei2022finetunedlanguagemodelszeroshot}. Pre SFT inference was done with grammar to enforce output structure.  Llama models were selected for SFT because of their accessibility and widespread adoption in clinical informatics research \citep{Nowak2025}. All fine-tuning and inference was performed on a HIPAA-secure virtual machine equipped with an A100 40GB GPU.

\subsection{Fine-Tuning}

We performed supervised fine-tuning of the LLMs using the Unsloth library \citep{unsloth}. The models were trained using rank-stabilized LoRA \citep{kalajdzievski2023rankstabilizationscalingfactor}, a parameter-efficient fine-tuning method that improves on the popular LoRA algorithm \citep{hu2021loralowrankadaptationlarge} and showed better performance in our experiments. The training parameters were found through initial exploration: rsLoRA rank and alpha of 16, 5 epochs, batch size of 2 and learning rate of 2e-4. The fine-tuning dataset corresponded to the notes from the breast cancer center ($n=487$), with no patient overlap with our test set. The fine-tuning process took one hour with the largest model (Llama 3.1 8B) and twenty minutes with the smallest model (Llama 3.2 1B).

\subsection{Postprocessing}

\begin{figure*}[!ht]
    \centering
    \includegraphics[width=0.95\textwidth]{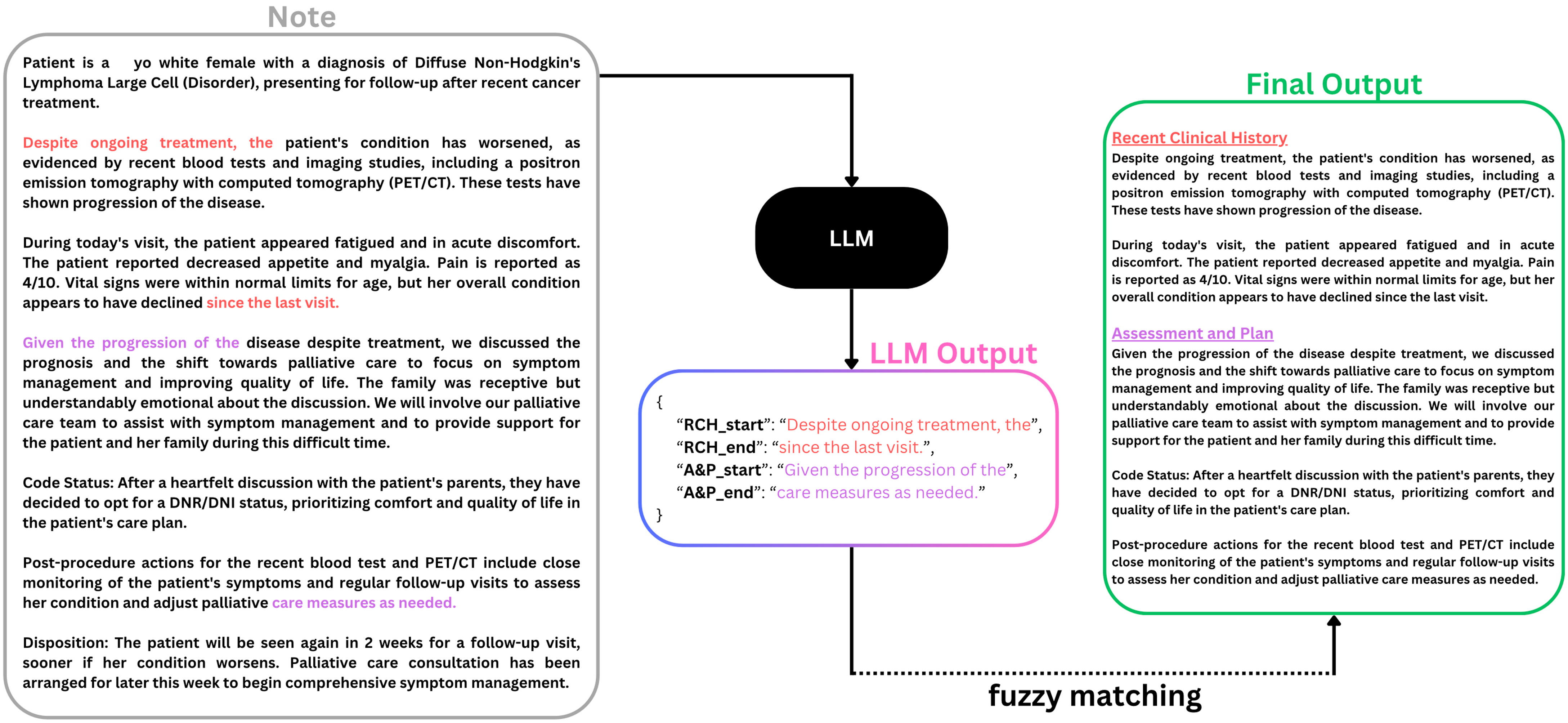}
    \caption{Section extraction workflow}
    \label{fig:sectioning-workflow}
\end{figure*}

An evaluation pipeline was implemented to process model outputs for each section of interest. Using vLLM \citep{vllm} to perform inference, the model was prompted to generate the first five words and the last five words of each predicted span \citep{Zhou2024}. These 5-grams were compared to the source text to identify matches. If a match was found, the segment from the identified starting position to the identified ending position was extracted and labeled as the 'predicted output' (\autoref{fig:sectioning-workflow}).\\

Due to the generative nature of LLMs, achieving an exact 5-gram match was uncommon, as observed in prior studies and in our experience \citep{Zhou2024}. To address this, fuzzy matching was employed to align the predicted start and end strings with the source text. This process used a sliding window of 5-grams derived from the source text and assessed similarity using the Levenshtein distance \citep{Levenshtein1966}, which measures the minimal number of edits required to transform one string into another. Matches with a similarity score exceeding 80\% were considered valid, ensuring robust identification of spans in the generated output that closely align with the source text.

\subsection{Evaluation}

\begin{figure}[H]
    \centering
    \includegraphics[width=0.95\linewidth]{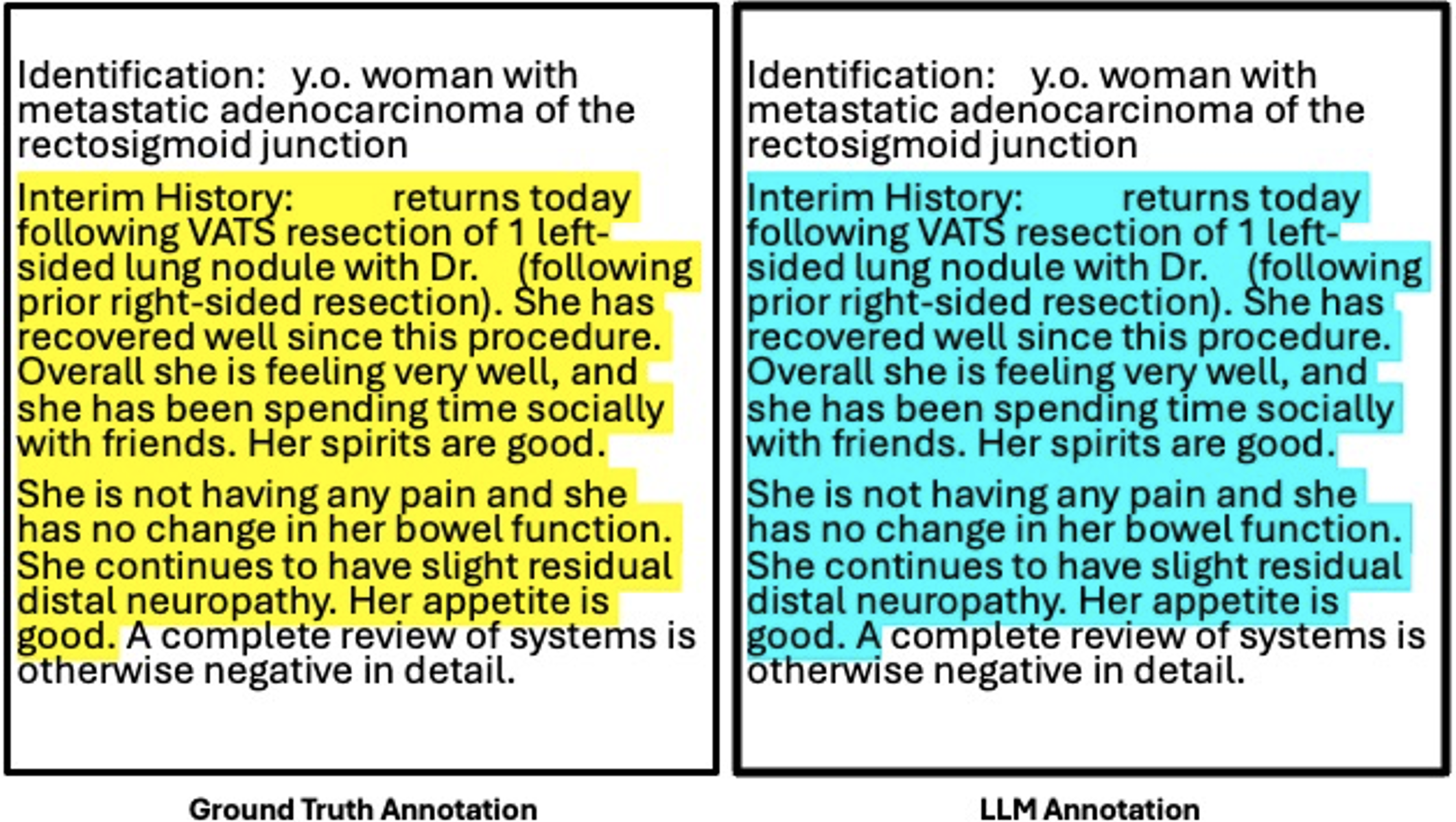}
    \caption{Labeled spans for RCH}
    \label{fig:labeled-rch}
\end{figure}

The predicted outputs were compared to ground truth annotations (\autoref{fig:labeled-rch}), and precision, recall, and F1 score were calculated. To assess model performance, we first ran inference three times on each model, then bootstrapped ($n=1,000$) each run to obtain 3,000 sets of metrics for evaluation. Statistical significance was assessed using a Friedman test ($\alpha=0.05$) \citep{Zimmerman1993}, with post-hoc pairwise comparisons via the Wilcoxon signed-rank test and a Bonferroni adjusted alpha of 0.01 \citep{Woolson2005WilcoxonST, Bland1995}.

\subsubsection{Internal Validity}

Outputs were evaluated on notes from two cancer centers (gastrointestinal and neurological), distinct from the cancer center used for training (breast), to assess performance across different patient populations at one institution. 

\subsubsection{External Validity}

To evaluate the external validity of this method, the best-performing model was used to section 50 progress notes from breast cancer patients at UCSF \citep{doi:10.1056/AIdbp2300110}. To ensure label consistency each note was annotated by KS using the validated codebook, and F1 scores were calculated.
\section{Results}

SecTag achieved an F1 score of 0.30 on the A\&P section but was unable to generate a valid output for the RCH section. The average F1 scores across both labels for MedSpaCy and Clinical-Longformer were 0.19 and 0.62, respectively. Detailed results of each approach can be found in \autoref{Baseline}.\\

\begin{table*}[hb!]
\centering
\setlength\aboverulesep{0pt}\setlength\belowrulesep{0pt}
\resizebox{1\linewidth}{!}{%
\begin{tabular}{l || c c | c c || c c | c c | c c } 
\toprule[1.5pt]
\multicolumn{1}{c ||}{\cellcolor[HTML]{EFEFEF}Model}
& \multicolumn{2}{c |}{\cellcolor[HTML]{EFEFEF}GPT-4o mini}
& \multicolumn{2}{c ||}{\cellcolor[HTML]{EFEFEF}GPT-4o}
& \multicolumn{2}{c |}{\cellcolor[HTML]{EFEFEF}\begin{tabular}[c]{@{}c@{}}Llama 3.2 1B \\{Instruct (FT)}\end{tabular}}
& \multicolumn{2}{c |}{\cellcolor[HTML]{EFEFEF}\begin{tabular}[c]{@{}c@{}}Llama 3.2 3B \\{Instruct (FT)}\end{tabular}}
& \multicolumn{2}{c }{\cellcolor[HTML]{EFEFEF}\begin{tabular}[c]{@{}c@{}}Llama 3.1 8B \\{Instruct (FT)}\end{tabular}}
\\
& RCH
& A\&P
& RCH
& A\&P
& RCH
& A\&P
& RCH
& A\&P
& RCH
& A\&P
\\
\midrule
F1 Score (95\% CI)
& \begin{tabular}[c]{@{}c@{}} 0.68 \\{(0.65-0.71)}\end{tabular}
& \begin{tabular}[c]{@{}c@{}} 0.72 \\{(0.69-0.74)}\end{tabular}
& \begin{tabular}[c]{@{}c@{}} 0.78 \\{(0.75-0.81)}\end{tabular}
& \begin{tabular}[c]{@{}c@{}} 0.79 \\{(0.77-0.82)}\end{tabular}
& \begin{tabular}[c]{@{}c@{}} 0.81 \\{(0.78-0.83)}\end{tabular}
& \begin{tabular}[c]{@{}c@{}} 0.90 \\{(0.88-0.92)}\end{tabular}
& \begin{tabular}[c]{@{}c@{}} 0.88 \\{(0.86-0.90)}\end{tabular}
& \begin{tabular}[c]{@{}c@{}} 0.92 \\{(0.90-0.93)}\end{tabular}
& \begin{tabular}[c]{@{}c@{}} 0.89 \\{(0.87-0.91)}\end{tabular}
& \begin{tabular}[c]{@{}c@{}} 0.94 \\{(0.93-0.95)}\end{tabular}
\\
Precision (95\% CI)
& \begin{tabular}[c]{@{}c@{}} 0.69 \\{(0.66-0.73)}\end{tabular}
& \begin{tabular}[c]{@{}c@{}} 0.72 \\{(0.69-0.75)}\end{tabular}
& \begin{tabular}[c]{@{}c@{}} 0.78 \\{(0.75-0.81)}\end{tabular}
& \begin{tabular}[c]{@{}c@{}} 0.79 \\{(0.76-0.81)}\end{tabular}
& \begin{tabular}[c]{@{}c@{}} 0.82 \\{(0.79-0.85)}\end{tabular}
& \begin{tabular}[c]{@{}c@{}} 0.91 \\{(0.89-0.93)}\end{tabular}
& \begin{tabular}[c]{@{}c@{}} 0.90 \\{(0.88-0.92)}\end{tabular}
& \begin{tabular}[c]{@{}c@{}} 0.94 \\{(0.92-0.95)}\end{tabular}
& \begin{tabular}[c]{@{}c@{}} 0.90 \\{(0.89-0.92)}\end{tabular}
& \begin{tabular}[c]{@{}c@{}} 0.94 \\{(0.93-0.96)}\end{tabular}
\\
Recall (95\% CI)
& \begin{tabular}[c]{@{}c@{}} 0.80 \\{(0.77-0.82)}\end{tabular}
& \begin{tabular}[c]{@{}c@{}} 0.86 \\{(0.84-0.88)}\end{tabular}
& \begin{tabular}[c]{@{}c@{}} 0.86 \\{(0.83-0.88)}\end{tabular}
& \begin{tabular}[c]{@{}c@{}} 0.88 \\{(0.86-0.90)}\end{tabular}
& \begin{tabular}[c]{@{}c@{}} 0.85 \\{(0.83-0.88)}\end{tabular}
& \begin{tabular}[c]{@{}c@{}} 0.92 \\{(0.90-0.94)}\end{tabular}
& \begin{tabular}[c]{@{}c@{}} 0.90 \\{(0.89-0.92)}\end{tabular}
& \begin{tabular}[c]{@{}c@{}} 0.91 \\{(0.90-0.93)}\end{tabular}
& \begin{tabular}[c]{@{}c@{}} 0.91 \\{(0.90-0.93)}\end{tabular}
& \begin{tabular}[c]{@{}c@{}} 0.95 \\{(0.94-0.96)}\end{tabular}
\\
\bottomrule[1.5pt]
\end{tabular}
}
\caption{Average performance of LLMs with 95\% confidence intervals}
\label{table-model-perf}
\end{table*}

We found that using SFT, the open source LLMs generated higher quality outputs relative to their base counterpart without the need for enforced structure (base model performance can be found in \autoref{base models}). Llama 3.1 8B had F1 scores of 0.89 and 0.94 for RCH and A\&P respectively (\autoref{table-model-perf}). The difference in model performance was statistically significant (p<0.01). Notably, Llama 3.1 8B scored 9-16 points higher than GPT-4o.\\

Error analysis was conducted on the top-performing model, Llama 3.1 8B, focusing on instances where the F1 score for a section fell below 0.8 (Gastrointestinal $n=96$, Neurological $n=24$). The most common error was over/underprediction of target section; detailed error analysis can be found in \autoref{Error analysis}.\\

On the 50 external progress notes, the F1 scores for RCH and A\&P using Llama 3.1 8B were 0.82 and 0.87 respectively.
\section{Discussion}

This study demonstrates that small, fine-tuned language models can outperform proprietary models in clinical section segmentation, offering significant advantages in cost, accuracy, and accessibility. Unlike proprietary models requiring institutional agreements and high computational costs \citep{Umeton2023}, our approach enables deployment on local or cloud-based systems, making it usable by researchers operating under resource constraints. This adaptability is crucial for downstream tasks such as symptom analysis and cohort discovery, where high-quality, actionable insights are critical.\\

Our findings demonstrate the potential of fine-tuning models with small datasets (fewer than 500 notes) to effectively perform note sectioning, even in the face of variability across clinical notes from different patient populations, offering a robust and adaptable solution for institutional use. Testing on notes from two distinct cancer populations and the progress notes of another institution highlights this approach's internal and external validity. While our study focused on progress notes, the strong performance demonstrates that fine-tuned models may effectively adapt to variations in note structure and content across institutions.\\

By integrating note sectioning with a small language model as a preprocessing step, the input size for larger, more resource-intensive language models in downstream tasks is significantly reduced. This reduction in input size decreases computational demands, leading to lower energy consumption and, consequently, a reduced carbon footprint \citep{stojkovic2024greenerllmsbringingenergyefficiency}. This approach underscores the potential for sustainable AI practices in clinical data processing by optimizing resource usage without compromising performance.\\

By providing a cost-effective and privacy-conscious solution, this work reduces reliance on proprietary systems. The affordability and accessibility of our approach ensures that high-quality research is no longer limited to large institutions, fostering innovation across diverse settings.\\

\textbf{Limitations}\\
While the model demonstrated strong performance overall, error analysis revealed patterns of overprediction and underprediction, particularly in sections with ambiguous or inconsistent boundaries. These errors highlight challenges posed by variability in clinical note structures and suggest areas for improvement, such as incorporating additional section labels to enhance discriminatory power. A potential mitigation strategy is incorporating a human-in-the-loop step to ensure sectioning aligns with study standards \citep{Chandler2022}.\\

This study focused exclusively on notes authored by physicians, nurse practitioners, and physician assistants, without evaluating notes written by other clinical staff, such as physical therapists, occupational therapists, or nutritionists. Furthermore, all analyzed notes originated from academic medical centers, limiting the assessment of variability in note styles across different types of hospital systems, such as community hospitals.
\section{Conclusion}

Our method demonstrates a robust, institution-agnostic solution for segmentation of clinical notes. By leveraging fine-tuned models that are cost-effective and adaptable, this approach offers a scalable and accessible methodology for improving clinical documentation analysis across diverse healthcare settings.

\section*{Conflicts of interest}
The authors have no competing interest to share.

\section*{Data availability}
The code used for this project as well as sample annotations based on the CORAL dataset are available in the following repository: \href{https://github.com/lindvalllab/MedSlice}{https://github.com/lindvalllab/MedSlice}

\bibliography{anthology,custom}
\bibliographystyle{acl_natbib}

\clearpage
\appendix

\onecolumn

\section{Annotation Codebook}
\label{Codebook}

\vspace{0.5cm}


\subsection*{General Guidelines}
In general, stick with annotating in big chunks rather than separated sections. It’s not possible if HPI and Interval History are separated by a large chunk of the oncology history, and that’s okay.\\

\subsection*{Recent Clinical History (HPI / Interval History)}

\textbf{Include:}
\begin{itemize}
    \item Anything in the following section heads/content:
    \begin{itemize}
        \item Interval History, interval treatment
        \item Subjective
        \item HPI, even if there is a lot of past Onc info in it, unless there is a separate section labeled Onc hx (then can omit that).
    \end{itemize}
    \item Free-hand documentation (e.g., unstructured communication notes with a patient at the bedside/clinic) that appear to be written without template.
    \item Text which looks like past interval history or past HPI but is not clearly demarcated by either a title (“HPI”), phrases, or another indication.
    \item Talk of a list of current symptoms that is outside the standard ROS and can clearly be seen as free-text documentation from the encounter.
    \begin{itemize}
        \item Example: “no nausea or itching.”
    \end{itemize}
\end{itemize}
\textbf{Exclude:}
\begin{itemize}
    \item The following sections: (even if they have something that might look important as it will be discussed again later on)
    \begin{itemize}
        \item Chief complaint
        \item Patient ID
        \item Reason for visit – UNLESS the words in there are the HPI!!
        \item Oncology history
        \item Review of systems
        \item Current treatment/therapy
        \item Templated lists of ESYM responses
        \item Patient instructions
        \item Referral orders
    \end{itemize}
    \item Information that is clearly copied forward, typically starts with or is followed by one of these sentences: Copied from, Above is for reference only, For reference, Carried through for continuity, Above history is for clinical reference only, Oncology history overview, OncHx has been copied forward and edited/updated from prior documentation for the purpose of clinical reference only, Oncology History, PMH, FH, and SH copied forward from previous notes and updated, included for clinical reference only.
\end{itemize}

\subsection*{Assessment and Plan}

\textbf{Include:}
\begin{itemize}
    \item Beginning at assessment and ending at the end of the follow-up instructions.
    \item Attending attestations (continue the same block of labeled text even if you include some things you normally would not).
    \item Statements about follow-up timing if it seems to be free text or there are clinical implications or information present.
    \item “IMP” = impression
    \item “Impression and recommendations”
\end{itemize}

\textbf{Exclude:}
\begin{itemize}
    \item Information that is copied forward: “Last assessment and plan.”
    \item Billing statements.
    \item “Verbalized understanding, all questions answered, will call…” unless it has non-templated writing like “for worsening pain.”
    \item Attestations if there is no free-written text, and it is just templated language, e.g., “I agree with assessment and plan with PA above.”
\end{itemize}


\newpage

\section{Prompt used for all LLMs}
\label{prompt}

\vspace{2mm}

\begin{quote}
    \textit{\textbf{Prompt:} Your task is to find the parts of a clinical note corresponding to the sections -History of Present Illness and Interval History-, and -Assessment and Plan-. You should organize this information in a JSON output that extracts the first and last five words for each of these sections. If the sections HPI\_Interval\_Hx or A\&P are not in the medical note, return an empty string for the corresponding section's start and end. Below is the medical note:}
\end{quote}

\vspace{1cm}

\section{Evaluation of sectioning approaches found in the litterature}
\label{Baseline}

\begin{table*}[ht!]
\centering
\setlength\aboverulesep{0pt}\setlength\belowrulesep{0pt}
\resizebox{0.55\linewidth}{!}{%
\begin{tabular}{l || c c | c c | c c } 
\toprule[1.5pt]
\multicolumn{1}{c ||}{\cellcolor[HTML]{EFEFEF}Model}
& \multicolumn{2}{c |}{\cellcolor[HTML]{EFEFEF}SecTag}
& \multicolumn{2}{c |}{\cellcolor[HTML]{EFEFEF}MedSpaCy}
& \multicolumn{2}{c }{\cellcolor[HTML]{EFEFEF}Clinical-Longformer}
\\
& RCH
& A\&P
& RCH
& A\&P
& RCH
& A\&P
\\
\midrule
F1 Score
& - 
& 0.30 
& 0.21 
& 0.16 
& 0.81 
& 0.63 
\\
Precision
& - 
& 0.31 
& 0.24 
& 0.16 
& 0.82 
& 0.64 
\\
Recall
& - 
& 0.42 
& 0.21 
& 0.16 
& 0.84 
& 0.65 
\\
\bottomrule[1.5pt]
\end{tabular}
}
\caption*{Average performance}
\label{table-baseline}
\end{table*}

\vspace{1cm}

\section{Performance of base models}
\label{base models}

\begin{table*}[h]
\centering
\setlength\aboverulesep{0pt}\setlength\belowrulesep{0pt}
\resizebox{0.8\linewidth}{!}{%
\begin{tabular}{l || c c | c c | c c } 
\toprule[1.5pt]
\multicolumn{1}{c ||}{\cellcolor[HTML]{EFEFEF}Model}
& \multicolumn{2}{c |}{\cellcolor[HTML]{EFEFEF}\begin{tabular}[c]{@{}c@{}}Llama 3.2 1B \\{Base}\end{tabular}}
& \multicolumn{2}{c |}{\cellcolor[HTML]{EFEFEF}\begin{tabular}[c]{@{}c@{}}Llama 3.2 3B \\{Base}\end{tabular}}
& \multicolumn{2}{c }{\cellcolor[HTML]{EFEFEF}\begin{tabular}[c]{@{}c@{}}Llama 3.1 8B \\{Base}\end{tabular}}
\\
& RCH
& A\&P
& RCH
& A\&P
& RCH
& A\&P
\\
\midrule
F1 Score (95\% CI)
& \begin{tabular}[c]{@{}c@{}} 0.14 \\{(0.12-0.16)}\end{tabular}
& \begin{tabular}[c]{@{}c@{}} 0.11 \\{(0.09-0.13)}\end{tabular}
& \begin{tabular}[c]{@{}c@{}} 0.35 \\{(0.32-0.38)}\end{tabular}
& \begin{tabular}[c]{@{}c@{}} 0.45 \\{(0.42-0.48)}\end{tabular}
& \begin{tabular}[c]{@{}c@{}} 0.53 \\{(0.50-0.55)}\end{tabular}
& \begin{tabular}[c]{@{}c@{}} 0.51 \\{(0.48-0.54)}\end{tabular}
\\
Precision (95\% CI)
& \begin{tabular}[c]{@{}c@{}} 0.21 \\{(0.18-0.24)}\end{tabular}
& \begin{tabular}[c]{@{}c@{}} 0.12 \\{(0.09-0.14)}\end{tabular}
& \begin{tabular}[c]{@{}c@{}} 0.52 \\{(0.49-0.56)}\end{tabular}
& \begin{tabular}[c]{@{}c@{}} 0.55 \\{(0.52-0.59)}\end{tabular}
& \begin{tabular}[c]{@{}c@{}} 0.69 \\{(0.66-0.73)}\end{tabular}
& \begin{tabular}[c]{@{}c@{}} 0.54 \\{(0.50-0.57)}\end{tabular}
\\
Recall (95\% CI)
& \begin{tabular}[c]{@{}c@{}} 0.14 \\{(0.11-0.16)}\end{tabular}
& \begin{tabular}[c]{@{}c@{}} 0.11 \\{(0.09-0.13)}\end{tabular}
& \begin{tabular}[c]{@{}c@{}} 0.40 \\{(0.37-0.43)}\end{tabular}
& \begin{tabular}[c]{@{}c@{}} 0.54 \\{(0.51-0.57)}\end{tabular}
& \begin{tabular}[c]{@{}c@{}} 0.51 \\{(0.48-0.53)}\end{tabular}
& \begin{tabular}[c]{@{}c@{}} 0.68 \\{(0.65-0.70)}\end{tabular}
\\
\bottomrule[1.5pt]
\end{tabular}
}
\caption*{Average performance of LLMs with 95\% confidence intervals}
\label{table-base-models}
\end{table*}

\newpage

\section{Error Analysis of Llama 3.1 8B Instruct on Gastrointestinal and Neurological Notes}
\label{Error analysis}

\begin{table*}[ht!]
\centering
\resizebox{\linewidth}{!}{%
\begin{tabular}{l|l|p{5cm}|p{6cm}|c|p{3cm}}
\toprule[1.5pt]
\textbf{Section} & \textbf{Type of Error} & \textbf{Description} & \textbf{Example} & \textbf{Count (\# Instances)} & \textbf{Details} \\ 
\bottomrule
RCH & 1. Slight over/underprediction & Negligible error; finds correct section & LLM includes a few extra characters at the end; LLM leaves out last 4 words (not important to the meaning of the sentence) & 3 & 1 over, 2 under \\ 
\cline{2-6}
    & 2. Moderate over/underprediction & Light error; finds correct section but includes too much preceding context under the section header; doesn't change meaning/readability of RCH & LLM includes introduction (e.g., "we had the pleasure of seeing…") prior to target paragraphs; LLM includes preceding paragraphs (intro) + includes extra 1.5 sentences (representing negative ROS) & 10 & 8 over, 2 both over and under \\ 
\cline{2-6}
    & 3. Notable over/underprediction & LLM either includes far too much text beyond the section or finds some of the correct section but reports incorrectly & LLM includes whole preceding paragraph, misses second sentence highlighted by human; LLM includes preceding "history of present illness (from previous note)" and also misses last sentence highlighted by human & 5 & 2 over, 3 both over and under \\ 
\cline{2-6}
    & 4. Hallucination of index & LLM produces an index that does not exist & Examples include hallucinated end index phrases like "on decadron No current facility-administered" or "neurologic deficits. Overall he has" that do not appear anywhere in the note & 3 & All 3 were errors with end index \\ 
\cline{2-6}
    & No prediction & No prediction; unsure whether this is a generation error, as RCH is not present in the note & N/A & 1 & \\ 
\cline{2-6}
    & No error & Error occurred for this record only regarding the A\&P section & N/A & 2 & \\ 
\bottomrule
A\&P & 1. Slight over/underprediction & Negligible error; finds correct section & Last sentence is "They know to contact me with any questions or concerns before their next visit."; LLM misses "before their next visit"; LLM included extra sentence: "he will call with any problems" & 10 & 7 over, 3 under \\ 
\cline{2-6}
    & 2. Moderate over/underprediction & Light error; finds correct section but includes too much preceding context under the section header; doesn't change meaning/readability of A\&P much & A\&P was one sentence, but LLM included this afterwards: "Please do not hesitate to contact me with any questions. I remain very interested in participating in the care of any" (weirdly cut off but doesn’t change the meaning of the captured information) & 5 & 4 over, 1 under \\ 
\cline{2-6}
    & 3. Notable over/underprediction & LLM either includes far too much text beyond the section or finds some of the correct section but reports incorrectly & LLM misses end of sentence, which could change meaning/readability: last sentence is "I will see her again in 1 year unless she has evidence of worsening cardiac function on surveillance echocardiography or symptoms of heart failure" \& LLM only captures up to "I will see her again in 1 year unless she has evidence of worsening cardiac function on" & 3 & 1 over, 2 under \\ 
\cline{2-6}
    & 3a. Possible error of processing? & LLM excludes character preceding a word (unsure whether this is an error) & Section reads "ASSESSMENT AND PLAN:? ?Hospitalization within" in note; LLM produces "ASSESSMENT AND PLAN:? Hospitalization within" & 1 & \\ 
\cline{2-6}
    & 4. Hallucination of index & LLM produces an index that does not exist & End index according to LLM: "continue to monitor clinically Plan:"; real end index: "Plan: - continue to monitor clinically" & 2 & Both were errors with end index \\ 
\cline{2-6}
    & 4a. Possible error of processing? & LLM may process space/characters weirdly (unsure whether this is an error) & Example: LLM correctly identifies general section but end index weird; unsure how "0- None ‚Ä¢ Skin Radiation" is processed. Note reads: "0- 

None

Skin  Radiation" & 1 & \\ 
\cline{2-6}
    & 5. Failed prediction and generation & LLM fails to identify section or generate & N/A & 1 & \\ 
\cline{2-6}
    & No error & No error found in this section & N/A & 3 & \\ 
\bottomrule[1.5pt]
\end{tabular}}
\caption*{Review of errors in the Neurological center}
\label{table-neuro-error-review}
\end{table*}

\begin{table*}[ht!]
\centering
\resizebox{\linewidth}{!}{%
\begin{tabular}{l|l|p{5cm}|p{6cm}|c|p{3cm}}
\toprule[1.5pt]
\textbf{Section} & \textbf{Type of Error} & \textbf{Description} & \textbf{Example} & \textbf{Count (\# Instances)} & \textbf{Details} \\ 
\bottomrule
RCH & 1. Slight over/underprediction & Negligible error; finds correct section & Same annotation but LLM adds extra word at the end & 31 & 12 slightly over, 6 over, 3 slightly under, 6 under, 4 both over and under \\ 
\cline{2-6}
    & 2. Notable over/underprediction & Error; finds some of the correct section but incorrectly reports & LLM includes ROS \& PMH; LLM misses language re: HPI preceding interval history & 32 & 19 under, 7 over, 6 both under and over \\ 
\cline{2-6}
    & 3. Failed prediction or generation & LLM fails to identify section or generate & For two notes, LLM correctly identifies section but fails to generate & 7 & 2 failed generation, 5 failed prediction + generation \\ 
\cline{2-6}
    & 4. Hallucination of index & LLM correctly identifies section but hallucinates end index & Correct end of section is "No bleeding" (this is followed by ROS) - LLM writes it twice "No bleeding No bleeding" - which is inaccurate & 4 & \\ 
\cline{2-6}
    & 5. Wrong section & LLM identifies the wrong section & N/A & 12 & \\ 
\cline{2-6}
    & 5a. No RCH in text; LLM finds something & No RCH present according to human annotation; LLM finds something & LLM finds medical history in absence of RCH & 8 & \\ 
\cline{2-6}
    & 5b. Misattributed section & LLM identifies real RCH as A\&P and picks something random for RCH & Random paragraph in chart review section & 1 & \\ 
\cline{2-6}
    & 5c. Picks wrong section & LLM identifies section titled 'HPI' or the like, but this is not the correct section & LLM found separate/repeat 'Interval History' section & 3 & \\ 
\cline{2-6}
    & No error & Complete overlap; error occurred only regarding A\&P or visually undetected spacing issue & N/A & 6 & \\ 
\cline{2-6}
    & Human error & Human annotation error & N/A & 3 & \\ 
\cline{2-6}
    & Human misnamed sections & Human annotated sections incorrectly & N/A & 2 & \\ 
\cline{2-6}
    & Human did not find section but LLM appeared to & Human failed to find section, but LLM captured it & N/A & 1 & \\ 
\bottomrule
A\&P & 1. Slight over/underprediction & Relatively negligible error; finds correct section; lack of final word may be confusing with underpredictions & "...and he is advised to start B12 1000 µg daily along with alpha lipoic acid 600 mg daily and a B complex..." (LLM misses the last word: "vitamin.") & 41 & 17 slightly under, 4 under, 23 slightly over, 5 over, 2 both slightly over and under, 1 both over and under \\ 
\cline{2-6}
    & 2. Notable over/underprediction & LLM includes too much information in A\&P section; misses potentially important parts of section & LLM includes preceding text within A\&P section; misses areas of A\&P section with multiple headers & 12 & 4 over, 6 under, 2 both over and under \\ 
\cline{2-6}
    & 3. Hallucination of index & LLM correctly identifies section but hallucinates end index & N/A & 5 & \\ 
\cline{2-6}
    & 4. Wrong section & No A\&P present; LLM finds RCH or random section & End index error as well in one of these (not counted in hallucination count - error attributed to main source (wrong section): reports "to be improved. Assessment: Sx appears" $\rightarrow$ this is not anywhere in the note. There is, however, "Assessment: Sx appears to be improved." & 4 & \\ 
\cline{2-6}
    & 5. Failed prediction or generation & LLM correctly identifies section but fails to generate OR LLM fails to identify section or generate & N/A & 7 & 2 failed generation, 5 failed prediction + generation \\ 
\cline{2-6}
    & No error & Error occurred for this record only regarding RCH section or visually undetected spacing issue & N/A & 12 & \\ 
\cline{2-6}
    & Human error & Human annotation error & N/A & 3 & \\ 
\cline{2-6}
    & Human misnamed sections & Human annotated sections incorrectly & N/A & 2 & \\ 
\cline{2-6}
    & Human did not find section but LLM appeared to & A\&P found by LLM; not found by human & N/A & 1 & \\ 
\cline{2-6}
    & Unsure & Unsure of whether error occurred; weird lines and spaces in start index may or may not be captured & "LLM captures 'Assessment \& Plan - Early satiety' for start index; however, this place in the note looks like 'Assessment \& Plan

-------------------------------  Early satiety -'. Unsure of whether the difference in these characters presents an error/issue or not." & 1 & \\ 
\bottomrule[1.5pt]
\end{tabular}}
\caption*{Review of errors in the Gastrointestinal center}
\label{table-gi-error-review}
\end{table*}

\end{document}